\documentclass[sigconf]{acmart}

\usepackage{booktabs}
\usepackage{multirow}
\usepackage{tabularx}
\usepackage[switch]{lineno}
\usepackage{enumitem}
\usepackage{graphicx} 
\setcopyright{acmcopyright}

\copyrightyear{2023}
\acmYear{2023}
\setcopyright{licensedothergov}\acmConference[SAC '23]{The 38th ACM/SIGAPP Symposium on Applied Computing}{March 27-March 31, 2023}{Tallinn, Estonia}
\acmBooktitle{The 38th ACM/SIGAPP Symposium on Applied Computing (SAC '23), March 27-March 31, 2023, Tallinn, Estonia}
\acmPrice{15.00}
\acmDOI{10.1145/3555776.3578577}
\acmISBN{978-1-4503-9517-5/23/03}

\begin{document}

\title{A Biomedical Entity Extraction Pipeline for\protect\\ Oncology Health Records in Portuguese}
  
\renewcommand{\shorttitle}{Biomedical Entity Extraction Pipeline for Oncology}

\author{Hugo Sousa}
\authornote{Both authors contributed equally to the paper}
\affiliation{\institution{INESC TEC, FCUP} \city{Porto} \state{Portugal}}
\email{hugo.o.sousa@inesctec.pt}

\author{Arian Pasquali}
\authornotemark[1]
\affiliation{\institution{INESC TEC} \city{Porto} \state{Portugal}}
\email{arianpasquali@gmail.com}

\author{Alípio Jorge}
\affiliation{\institution{INESC TEC, FCUP} \city{Porto} \state{Portugal}}
\email{alipio.jorge@inesctec.pt}

\author{Catarina Sousa Santos} 
\affiliation{ \institution{INESC TEC} \city{Porto} \state{Portugal}}
\email{catarina.s.santos@inesctec.pt}

\author{Mário Amorim Lopes}
\affiliation{\institution{INESC TEC} \city{Porto} \state{Portugal}}
\email{mario.a.lopes@inesctec.pt}

\renewcommand{\shortauthors}{H. Sousa and A. Pasquali et al.}

\begin{abstract}
Textual health records of cancer patients are usually protracted and highly unstructured, making it very time-consuming for health professionals to get a complete overview of the patient's therapeutic course. As such limitations can lead to suboptimal and/or inefficient treatment procedures, healthcare providers would greatly benefit from a system that effectively summarizes the information of those records. With the advent of deep neural models, this objective has been partially attained for English clinical texts, however, the research community still lacks an effective solution for languages with limited resources. In this paper, we present the approach we developed to extract procedures, drugs, and diseases from oncology health records written in European Portuguese. This project was conducted in collaboration with the Portuguese Institute for Oncology which, besides holding over $10$ years of duly protected medical records, also provided oncologist expertise throughout the development of the project.  Since there is no annotated corpus for biomedical entity extraction in Portuguese, we also present the strategy we followed in annotating the corpus for the development of the models. The final models, which combined a neural architecture with entity linking, achieved $F_1$ scores of $88.6$, $95.0$, and $55.8$ per cent in the mention extraction of procedures, drugs, and diseases, respectively.  
\end{abstract}


\maketitle
\section{Introduction}
\label{sec:intro}

The effectiveness achieved by artificial intelligence (AI) systems in recent years has led to an increase in practical applications in healthcare support. To give an example, the number of FDA-approved AI medical algorithms has grown rapidly in recent years \cite{Ebrahimian2022FDA-regulatedStudies}. Such growth would not be possible without the digitization of healthcare, which has significantly increased the number of clinical records stored in digital collections. These longitudinal records -- known as electronic health records (EHRs) -- compile all available patient-specific information, such as previous conditions, diagnoses, treatments, prescriptions, laboratory test results, radiological images, and phenotypic and genotypic features \cite{Jensen2012MiningCare,Bibault2016BigProspects}. As a result, much potential lurks within EHRs \cite{Wang2018ClinicalReview}. First, from an integrated healthcare standpoint, by providing clinicians with instant and portable access to all up-to-date patient information, EHRs streamline local and remote care delivery, potentially leading to better-informed decision-making \cite{Graber2017TheDiagnosis,Jensen2012MiningCare,Bibault2016BigProspects}. Second, the possible benefits of mining these records for research purposes are endless. Current research of AI methods in the clinical context includes: variant reporting \cite{Bibault2016BigProspects}; public health and environmental impact surveillance \cite{Boland2021HarnessingPFAS,Lee2018NaturalRecords}; and precision (or personalized) Medicine \cite{Fagherazzi2020DeepDeeper}.

Although most successful applications of AI in healthcare have been for image-based diagnosis and detection \cite{Liu2019AMeta-analysis}, most of EHRs are primarily composed of free-text clinical narratives describing the patient's history, which opens up a wide range of possibilities for natural language processing (NLP) applications. Nevertheless, due to their unstructured nature, harnessing the information within these records is not straightforward. When compared with a more structured representation, the free-text format has the advantage of allowing the medical staff to communicate more naturally, providing adequate freedom for clear human-to-human communication. On the other hand, since the amount of text grows with the patient's interactions with healthcare professionals, the documents quickly reach dozens and hundreds of pages. This problem is aggravated in patients with persistent diseases, such as cancer, where the number of pages can run into the thousands. Such lengthy documents are intractable for the clinical staff to handle, especially in an emergency setting where time is scarce to take action. In such a scenario, a tool that would automatically and reliably summarize the patient's therapeutic path would be a great asset for improving healthcare services. 

An important step to achieve such summarization is the identification/extraction of the biomedical entities in the text -- such as diseases, medical procedures, or drugs. Although some systems were developed for general-purpose clinical texts written in English -- including Apache cTAKES\footnote{\url{https://ctakes.apache.org/}}, CLAMP\footnote{\url{https://clamp.uth.edu/}}, and more notably IBM Watson Health\footnote{\url{https://www.ibm.com/watson-health}} --  the same can not be said for languages with relatively low-resources, where effective extraction of medical entities remains a challenge \cite{AlShuweihi2021BiomedicalReview}. The lack of effectiveness is mainly attributed to the scarcity or nonexistence of publicly available data, which limits the development of tools for clinical texts (such as the pre-training of large language models for the clinical setting). Apart from that, the heterogeneity of clinical narratives is also a major problem. For instance, while the clinical note that results from a scheduled appointment is descriptive and syntactically correct, the same is not true for notes from the emergency service where itemized text is prevalent and the use of abbreviations and acronyms more frequent. Therefore, a model that is trained in scheduled appointment documents would (most likely) present a much lower effectiveness in emergency service documents. In practice, this means that to achieve an effective system, one must annotate a corpus within the domain/institution where the model will be deployed. Since the current state-of-the-art systems (which are based on deep neural models) require large amounts of annotated data to attain high effectiveness, this is a major bottleneck in system development.

Faced with the aforementioned issues while developing a system for the extraction of biomedical entities for the Portuguese Institute for Oncology (IPO) in Porto, Portugal's largest oncology center, we created an efficient approach for the annotation of a medical corpus that can be extrapolated for any medical domain and institution to quickly achieve a corpus that can be used to train a deep neural model. In this paper, we describe how the Unified Medical Language System (UMLS) metathesaurus was employed to accelerate and minimize the required resources for the annotation process of the corpus. Furthermore, in this paper, we also show that by combining three named entity recognition models -- for the identification of procedures, drugs, and diseases -- with entity linking on UMLS we were able to achieve relevant effectiveness for European Portuguese. As this approach is only limited by the availability of a pre-trained large language model and the availability of the language on the UMLS knowledge base, this approach can also be extrapolated to other languages and domains. 

In summary, the main contributions of this work are the following:
\begin{enumerate}[label=(\roman*)]

\item we present an efficient strategy for the annotation of clinical narratives with biomedical entities that accelerates the annotation process;

\item we present a methodology to achieve efficient biomedical entity extraction for Portuguese that can be adapted to other languages with equivalent resources;

\item finally, we present a real-life case study with results obtained by applying such a strategy for the extraction of biomedical entities on free-text oncology narratives written in European Portuguese.

\end{enumerate}

The remainder of the paper is structured as follows: in Section \ref{sec:related_work} we frame the presented work in the context of the current research. Section \ref{sec:data} describes the developed methodology for the annotation of clinical texts along with the corpus that resulted from applying it in IPO Porto documents. In section \ref{sec:model} we present the prediction pipeline that was implemented on IPO Porto and present the experimental procedure that led us to that architecture in Section \ref{sec:experiments}. Finally, Section \ref{sec:conclusion} and \ref{sec:future} concludes the paper by discussing possibilities for future research, respectively.

\section{Related Work}
\label{sec:related_work}

The aim of biomedical NER (BioNER) \cite{Kim2019AMulti-Domains} is to identify the spans of biomedical entities in a clinical text. Not long ago, the best-performing systems relied on hand-crafted rules to capture different characteristics of entity classes \cite{Leser2005WhatLiterature,Campos2012BiomedicalTools}. However, these features frequently result in highly specialized tools that can only be used in the domains for which they were designed.  Recently, the developments in NLP, in particular with large language models such as BERT \cite{Vaswani2017AttentionNeed}, BioBERT \cite{Lee2020BioBERT:Mining}, or PubMedBERT \cite{Gu2021Domain-SpecificProcessing}, have mitigated this problem. Moreover, training deep learning architectures such as Long Short-Term Memory (LSTM) and Conditional Random Field (CRF) on top of the large language models word representations, had greatly improved the effectiveness of BioNER for English texts over the last few years \cite{Habibi2017DeepRecognition,Giorgi2018TransferNetworks,Wang2018Cross-typeLearning,Yoon2019CollaboNet:Recognition}. 

In that vein, a few deep-learning-based NLP approaches have been developed to support clinical decision-making in cancer-related contexts using unstructured clinical texts. These have been tested in several medical fields and applications, such as but not limited to: senology, to identify and pinpoint local and distant recurrence for breast cancer \cite{Karimi2021DevelopmentData}; endocrinology, to improve the diagnosis of thyroid cancer \cite{Zhang2022ImprovedStudy}; urology, to stratify the risk of developing urinary incontinence following partial or full prostate removal \cite{Bozkurt2020PhenotypingCase}; and clinical trial screenings, to select patients based on prognosis and propensity to changes in therapeutic courses \cite{Kehl2021ClinicalCancer}. Additionally, a BERT-based model -- MetBERT  -- was also created for the prediction of metastatic cancer from clinical notes \cite{Liu2022MetBERT:Notes}.

Nevertheless, developing effective BioNER models for low-resource languages remains a challenge \cite{Neveol2018ClinicalChallenges,Lerner2020TerminologiesRecognition}. This is primarily because harnessing the power of deep learning models requires a substantial amount of data. However, because clinical texts contain highly sensitive information, ethical and legal constraints make it hard to make corpora available. Yet another issue is the specificity of the data, which limits its applicability.  For example, the only clinical text corpus released for European Portuguese \cite{Lopes2019ContributionsPortuguese} does not resemble the clinical notes we found at IPO, as the latter were more disaggregated/itemized and did not follow the traditional semantic structure of a narrative. Consequently, application projects are forced to develop a corpus for the data with which the application will go into production. In this regard, some efforts have been made to develop methodologies to simplify the annotation of the clinical corpus. For instance, in a similar challenge to ours, \citet{Silvestri2022IterativeBases} used active learning and distant supervision to annotate clinical texts written in Italian. In this work, we adopted a more general strategy that requires less computational power.
\section{Corpus Development}
\label{sec:data}

IPO Porto has been keeping electronic health records of its patients for more than $10$ years. We were able to work with a sufficient sample, under confidentiality, and on-site. Among the many files that describe each patient's journey through the hospital, there is one of particular interest for this project, the clinical diary. 


The clinical diary contains a textual description of every action or interaction the patient had in the hospital. This includes any treatment, procedure, appointment, plan of action, and other information deemed relevant by the health professionals. It is a very descriptive file and, thereby, a rich source of information. Adversely, it can become very long, repetitive, and hard to be efficiently read by humans in the loop.

\subsection{Sampling Process}

For the development of the corpus, we selected $80$ fairly short clinic diaries. In practical terms, this meant sampling documents that were between the first and second quartile in terms of the number of pages, which materialized in documents ranging from $52$ to $97$ A4 equivalent pages long. This is significant because we wanted to reflect a rich variety of clinical note-writing styles as possible in the corpus (which vary with the person that writes them). Yet another benefit of sampling from a variety of relatively small documents is that it increases the exposure to different patients. Consequently, this increases the number of biomedical entities that will be present in the final corpus. 

\subsection{Preprocessing}

The clinical diary is composed of clinical notes that are delimited by headers. Important metadata, including the date and the context in which the note was written, is contained in the headers. Since the status of the patient does not change in every interaction that he/she has with the medical staff, it is common that the clinical notes overlap in content. Furthermore, there is information (such as comorbidities and pathological antecedents) that is frequently repeated throughout the clinical diary. To avoid multiple annotations of the same content, we segmented the clinical notes into sentences with the \texttt{segtok} sentence segmenter\footnote{\url{https://github.com/fnl/segtok}}. From the $80$ clinic diaries, we collected a total of $36,512$ unique sentences.

Yet another important specificity of clinical notes is the frequent use of acronyms and abbreviations which are commonly employed by clinical staff. Although there has been some work towards using word embeddings/deep neural models to handle the acronyms and abbreviations \cite{Liu2015ExploitingExpansion,Li2019ADisambiguation} we found it sufficiently simple and robust to build a glossary of the acronyms and abbreviations found in our corpus. In the current version, the glossary contains $183$ entries. However, it can be manually extended to include new acronyms and abbreviations that may appear in future documents.

\subsection{Annotation with UMLS}

\begin{table}[t]
\centering
\caption{\label{tab:umls} How the UMLS semantic types were grouped for our three entity types. TUI stands for ``Type Unique Identifier''.}
\begin{tabularx}{\linewidth}{lll} 
\midrule
                            & \textbf{TUI} & \textbf{Name}                        \\ 
\midrule
\multirow{4}{*}{\textbf{Procedures}} & T058         & Health Care Activity                 \\
                            & T059         & Laboratory Procedure                 \\
                            & T060         & Diagnostic Procedure                 \\
                            & T061         & Therapeutic or Preventive Procedure  \\ 
\midrule
\multirow{4}{*}{\textbf{Drugs}}      & T121         & Pharmacologic Substance              \\
                            & T122         & Biomedical or Dental Material        \\
                            & T195         & Antibiotic                           \\
                            & T200         & Clinical Drug                        \\ 
\midrule
\multirow{12}{*}{\textbf{Diseases}}  & T019         & Congenital Abnormality               \\
                            & T020         & Acquired Abnormality                 \\
                            & T033         & Finding                              \\
                            & T037         & Injury or Poisoning                  \\
                            & T046         & Pathologic Function                  \\
                            & T047         & Disease or Syndrome                  \\
                            & T048         & Mental or Behavioral Dysfunction     \\
                            & T049         & Cell or Molecular Dysfunction        \\
                            & T050         & Experimental Model of Disease        \\
                            & T184         & Sign or Symptom                      \\
                            & T190         & Anatomical Abnormality               \\
                            & T191         & Neoplastic Process                   \\
\bottomrule
\end{tabularx}
\end{table}

\begin{table*}[t]
\centering
\caption{\label{tab:data} Distribution of the corpus statistics over train, validation, and test sets within the three entity types (procedures, drugs, and diseases). \emph{Sent}, \emph{B}, \emph{I}, and \emph{O} stand for the number of sentences, begin tokens, inside tokens, and outside tokens, respectively.}
\resizebox{\linewidth}{!}{%
\begin{tabular}{llrrrrlrrrrlrrrrlrrrr} 
\cmidrule[\heavyrulewidth]{3-21}
\multicolumn{1}{c}{} & \multicolumn{1}{c}{} & \multicolumn{4}{c}{\textbf{Procedures}}                                                          & \multicolumn{1}{c}{} & \multicolumn{4}{c}{\textbf{Drugs}}                                                               & \multicolumn{1}{c}{} & \multicolumn{4}{c}{\textbf{Diseases}}                                                            & \multicolumn{1}{c}{} & \multicolumn{4}{c}{\textbf{Aggregated}}                                                           \\ 
\cmidrule{3-21}
\multicolumn{1}{c}{} & \multicolumn{1}{c}{} & \multicolumn{1}{c}{Sent} & \multicolumn{1}{c}{B} & \multicolumn{1}{c}{I} & \multicolumn{1}{c}{O} & \multicolumn{1}{c}{} & \multicolumn{1}{c}{Sent} & \multicolumn{1}{c}{B} & \multicolumn{1}{c}{I} & \multicolumn{1}{c}{O} & \multicolumn{1}{c}{} & \multicolumn{1}{c}{Sent} & \multicolumn{1}{c}{B} & \multicolumn{1}{c}{I} & \multicolumn{1}{c}{O} & \multicolumn{1}{c}{} & \multicolumn{1}{c}{Sent} & \multicolumn{1}{c}{B} & \multicolumn{1}{c}{I} & \multicolumn{1}{c}{O}  \\ 
\midrule
Train                &                      & 2,398                    & 2,091                 & 1,225                 & 39,333                &                      & 1,236                    & 1,128                 & 6                     & 16,529                &                      & 1,870                    & 1,651                 & 512                   & 26,616                &                      & 4,219                    & 4,870                 & 1,743                 & 62,328                 \\
Validation           &                      & 599                      & 610                   & 311                   & 8,717                 &                      & 309                      & 285                   & 2                     & 4,163                 &                      & 467                      & 395                   & 141                   & 5,738                 &                      & 1,055                    & 1,290                 & 454                   & 13,503                 \\
Test                 &                      & 701                      & 821                   & 452                   & 13,992                &                      & 377                      & 445                   & 1                     & 6,737                 &                      & 608                      & 734                   & 210                   & 11,743                &                      & 1,318                    & 2,000                 & 663                   & 25,755                 \\ 
\midrule
\textbf{Total}       &                      & 3,698                    & 3,522                 & 1,988                 & 62,042                &                      & 1,922                    & 1,858                 & 9                     & 27,429                &                      & 2,945                    & 2,780                 & 863                   & 44,097                &                      & 6,592                    & 8,160                 & 2,860                 & 101,586                \\
\bottomrule
\end{tabular}
}
\end{table*}

To greatly reduce manual annotation effort we start by automatically annotating the sentences. For this purpose, the Unified Medical Language System (UMLS) metathesaurus \cite{Bodenreider2004TheTerminology} was used as a knowledge base for biomedical entity identification. UMLS contains an ontology that maps each biomedical entity (referred to as ``concept'' in UMLS jargon) to a semantic group, which is a high-level categorization of the entities. Examples of semantic groups are ``Clinical Drug'', ``Pathologic Function'', and ``Diagnostic Procedure''. To determine the semantic groups of interest in our project, we followed a two-step process.  First, we asked the oncology team to pre-select the semantic groups to be extracted based on the description of the semantic group\footnote{The semantic groups and their respective definition can be found in \url{https://lhncbc.nlm.nih.gov/ii/tools/MetaMap/Docs/SemGroups_2018.txt}}. We then annotated 300 sentences with the entities within the initial set of semantic groups and presented the result of the annotation to the oncology team.  In conjunction, we parsed all sentences and iteratively updated the set of semantic groups to add concepts that were not annotated (that is, false negatives). The final set of semantic groups was then grouped into three types: procedures, drugs, and diseases. Table \ref{tab:umls} presents the mapping between semantic groups and the defined entity types\footnote{For the complete semantic type mappings see \url{https://lhncbc.nlm.nih.gov/ii/tools/MetaMap/documentation/SemanticTypesAndGroups.html}}. Our initial automatic labeling step aims to catch as many true entities as possible even if at the cost of including many false entities. By following this procedure, we ultimately produced a high-recall classifier, and therefore, reduce the annotation effort of the specialists to eliminate the false entities.

To extract the concepts that were within the final set of semantic groups, we used the QuickUMLS toolkit\footnote{\url{https://github.com/Georgetown-IR-Lab/QuickUMLS}} \cite{Soldaini2016QuickUMLS:Extraction}, which implements fast string matching with the UMLS concepts. From the set of $36,512$ sentences, QuickUMLS identified concepts in $9,912$ of them.

\subsection{Specialized Curation}

Since the set of semantic groups to be extracted by UMLS was selected to maximize recall, the annotation produced by UMLS  included a lot of vague terms (such as "blood" or "pain") that are not useful for oncologists. To curate the annotations, we presented the annotations to the oncology team and instructed them to validate the annotations, i.e., remove false positives. This is a significant boost in the manual annotation effort as it does not require the specialist to read the entire sentence, but only to look at the entities that were annotated by UMLS. However, the annotators were also allowed to add missing entities, in case they drew their attention while parsing the UMLS annotations. 

Note that, since the annotation produced by UMLS covers a wide range of concepts, the revision by the specialist can be interpreted as the selection of the entities he/she wants the pipeline to identify. Therefore, the resulting dataset will be tailored to the necessities and requirements of the institution. 

After the curation of the specialists and the removal of all the sentences that did not have any annotation, we were left with  $6,592$ sentences, in which, $3,689$ contained procedures mentions, $1,922$ contained drugs mentions, and $2,945$ contained diseases mentions. As the goal was to train a NER model, the annotation was stored in the Inside-Outside-Beginning format \cite{Ramshaw1995TextLearning}. The last row of Table \ref{tab:data} shows how the sentences and the begin/inside/outside tokens were distributed over the datasets that only contained one entity type (``Procedures'', ``Drugs'', and ``Diseases'' columns), and the dataset with all the entity types (``Aggregated'' column). The remaining rows of the table present the distribution over the train, validation, and test set. This split was performed randomly at the sentence level of the ``Aggregated'' dataset. First, $20\%$ of the sentences were set aside for test. Then, the remaining $80\%$ of sentences were partitioned so that $80\%$ of the documents were used to train the models and $20\%$ to validate them.

\section{Prediction Pipeline}
\label{sec:model}

\begin{figure*}
    \centering
    \includegraphics[width=\linewidth]{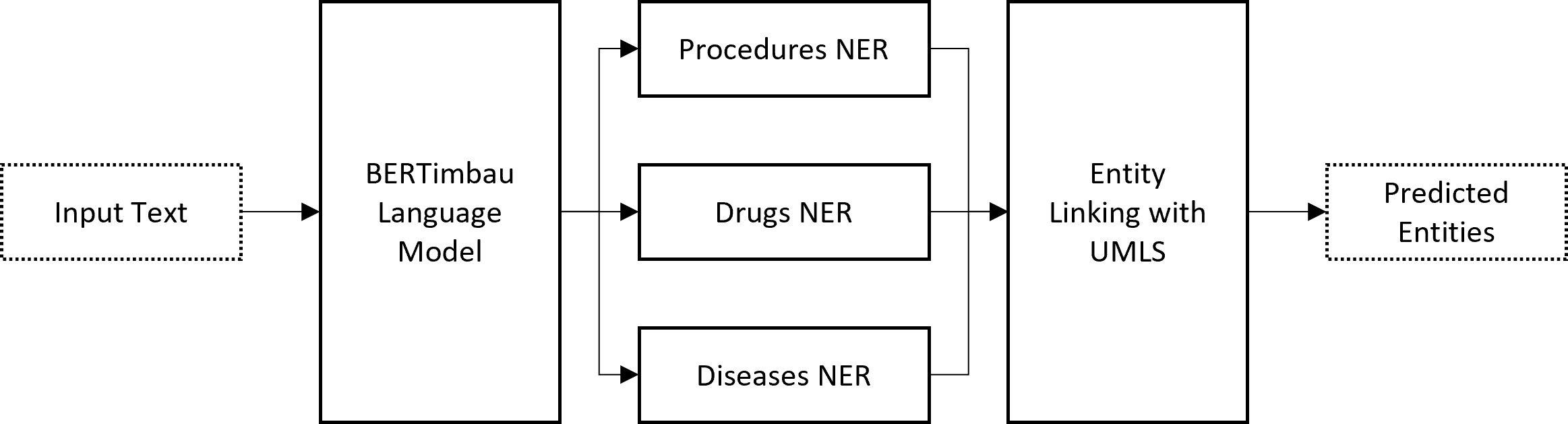} 
    \caption{Graphical representation of the proposed pipeline.}
    \label{fig:pipeline}
\end{figure*}

To model the biomedical entity identification task we designed a pipeline that conceptually can be segmented into three components: a pre-trained language model to provide embeddings for the input text; the NER model(s) to identify the entities of interest; and entity linking with UMLS to ensure that entities that were found by the NER models refer to biomedical entities. Figure \ref{fig:pipeline} presents a graphical illustration of the architecture we used on our specific application for IPO Porto. 

For the pre-trained language model, we employed the  BERTimbau\footnote{\url{https://huggingface.co/neuralmind/bert-base-portuguese-cased}} \cite{Souza2020BERTimbau:Portuguese} language model, which was built by training the BERT \cite{Devlin2019BERT:Understanding} architecture in a mixture of  Brazilian and European Portuguese texts\footnote{There is currently no large-language model trained exclusively on European Portuguese texts.}. For the identification of the entities, we trained three models -- one for each entity type -- using the architecture proposed by \citet{Ma2016End-to-endLSTM-CNNs-CRF}. This model architecture was tailored for sequence labeling, relying on a convolutional neural network \cite{LeCun1989BackpropagationRecognition} for character encoding, a bi-directional long-short term memory \cite{Hochreiter1997LongMemory} to aggregate character and word representations, and a sequential conditional random field \cite{Lafferty2001ConditionalData} to jointly classify all words of a sentence. Lastly, all the entities identified by the NER models are looked for in the UMLS (a process commonly referred to as entity linking) with the list of semantic types we deemed relevant in the automatic annotation phase (see Table \ref{tab:umls}). If the entity is found in the UMLS and its semantic group is part of the set that is relevant, the entity is kept, otherwise, it is discarded.

An important remark is that although we use a NER model for procedures, drugs, and diseases, the final class of the entity is defined by the metadata in the UMLS. More precisely, by its semantic group, which is then mapped to the class by the map in Table \ref{tab:umls}. Conceptually, the three NER models can be interpreted as just one NER model that classifies the tokens as \texttt{BEGIN}, \texttt{INSIDE}, and \texttt{OUTSIDE} rather than classifying the tokens as \texttt{B-PROCEDURE}, \texttt{B-DRUG}, et cetera.
\section{Experiments}
\label{sec:experiments}

\subsection{Evaluation}
For the evaluation and comparison of the models, we employed the strict and relaxed Precision, Recall, and $F_1$ metrics. Whereas the strict metrics only count exact matches, the relaxed metrics consider the overlaps between the predictions and the annotations as being correct. Therefore, by presenting both we are able to attain a more comprehensive understanding of the model's effectiveness.  

Naturally, for the dataset, we used the corpus that is described in Section \ref{sec:data}. To provide a detailed overview of the model's effectiveness, besides presenting the result on the Aggregated dataset (the dataset that contains the annotation of all the entity types) we also analyze the results of the model on the Procedures, Drugs, and Diseases datasets. The train, validation, and test partitions are present in Table \ref{tab:data}. 

\subsection{Models}
To evaluate the importance of the components in our prediction pipeline we run experiments with three modeling settings: just using the UMLS predictions; just using the NER predictions; and using the full pipeline prediction. Bellow follows a detailed description of the three approaches.

\paragraph{\textbf{UMLS}} For the baseline model we used UMLS as introduced in Section \ref{sec:data}. That is, we used the QuickUMLS tool to search for concepts belonging to the semantic groups that were deemed relevant for each entity type (see Table \ref{tab:umls}). This is a simple baseline that can be easily reproduced and therefore, provide a reference metrics that can serve as a base for comparison in future research.

\paragraph{\textbf{NER}} The joint NER task is approached by first aggregating the predictions of three separate, but structurally identical, NER models. One for identifying mentions to procedures, another for diseases and a third for drugs. This serves to evaluate the effectiveness of the NER models without doing entity linking with UMLS. As we use three different models, one token may be classified with two entity types. That is, a token can be classified as \texttt{B-DRUG} and \texttt{B-DISEASE}. In that case, we check the raw values outputted by the NER models and keep the entity with the highest value.

\paragraph{\textbf{NER \& UMLS}} Lastly, the NER \& UMLS is the model described in Section  \ref{sec:model} that does entity linking of the entities predicted by the NER model.

\subsection{Implementation Details}
The pipeline was implemented in Python. In particular, we used the QuickUMLS toolkit to implement the UMLS entity linking, and Spark NLP to train and implement the NER models. All the experiments were run on a server with an AMD EPYC 7351P processor and NVIDIA GeForce GTX 1080 Ti GPU.

\subsection{Training}

As stated in Section \ref{sec:model}, three NER models were trained for the identification of procedures, drugs, and diseases. Before training, we randomly sampled a set of $20$ hyperparameter configurations to train each of the models. The hyperparameters that were tuned are: the dropout rate, which was sampled from $\{0.0$, $0.1$, $0.2$, $0.3$, $0.4$, $0.5\}$; the batch size, which was sampled from $\{4$, $8$, $16$, $32$, $64$, $128\}$; and the learning rate, which was sampled from $\{0.001$, $0.003$, $0.005$, $0.01$, $0.03$, $0.05$, $0.1\}$. For each hyperparameter configuration, we trained the model for $30$ epochs and stored the weights of the epoch in which the model had the best micro $F_1$ score in the validation set. After training the $20$ models for each entity type, we kept the model with the best validation $F_1$ as our final model.

\subsection{Results}
\label{sec:results}

\begin{table*}[t]
\centering
\caption{\label{tab:results} Results of evaluating the three models on test partition of the datasets. In bold and underlined are highlighted the best strict and relaxed $F_1$, respectively.}
\resizebox{\linewidth}{!}{%
\begin{tabular}{llcccccclcccccclcccccc} 
\cmidrule[\heavyrulewidth]{3-22}
\multicolumn{1}{c}{} &                   & \multicolumn{6}{c}{\textbf{UMLS}}                         & \multicolumn{1}{c}{} & \multicolumn{6}{c}{\textbf{NER}}                         & \multicolumn{1}{c}{} & \multicolumn{6}{c}{\textbf{NER \& UMLS}}                      \\ 
\cmidrule{3-22}
\multicolumn{1}{c}{} &                   & \multicolumn{3}{c}{Strict}  & \multicolumn{3}{c}{Relaxed} &                      & \multicolumn{3}{c}{Strict} & \multicolumn{3}{c}{Relaxed} &                      & \multicolumn{3}{c}{Strict}  & \multicolumn{3}{c}{Relaxed}  \\
\multicolumn{1}{c}{} & ~ & P    & R    & F1            & P    & R    & F1            & ~    & P    & R    & F1           & P    & R    & F1            & ~    & P    & R    & F1            & P    & R    & F1             \\ 
\midrule
\textbf{Procedures}           &                   & 65.4 & 96.7 & 78.0          & 68.5 & 99.9 & 81.3          &                      & 75.2 & 80.8 & 77.9         & 81.0 & 87.0 & 83.9          &                      & 95.0 & 83.1 & \textbf{88.6} & 98.2 & 85.6 & \underline{91.4}           \\
\textbf{Drugs}                &                   & 75.8 & 92.5 & 83.3          & 76.2 & 93.0 & 83.7          &                      & 74.5 & 95.2 & 83.6         & 74.5 & 95.2 & 83.6          &                      & 98.6 & 91.6 & \textbf{95.0} & 98.6 & 91.6 & \underline{95.0}           \\
\textbf{Diseases}             &                   & 67.6 & 97.2 & \textbf{79.7} & 70.1 & 99.8 & \underline{82.4}          &                      & 57.4 & 46.0 & 51.1         & 87.1 & 69.7 & 77.4          &                      & 67.5 & 47.5 & 55.8          & 95.7 & 66.8 & 78.7           \\
\textbf{Aggregated}                  &                   & 44.5 & 97.0 & 61.0          & 47.3 & 99.7 & 64.2          &                      & 62.3 & 68.0 & 65.0         & 75.7 & 81.8 & 78.6          &                      & 79.8 & 67.3 & \textbf{73.0} & 92.3 & 77.4 & \underline{84.2}           \\
\bottomrule
\end{tabular}
}
\end{table*}

The results of the final models on the test set of the datasets are presented in Table \ref{tab:results}. In a general overview of the effectiveness of the three models, one can observe that the NER \& UMLS model presents the best strict and relaxed $F_1$ in all the datasets except for the Diseases, where the UMLS model presents the best results. However note that while the strict $F_1$ of the NER \& UMLS model for the identification of Diseases is 23.9 percentage points worst than the UMLS model, when we compare the relaxed $F_1$ score the difference is only 3.7 percentage points. Upon close inspection, we found that UMLS has a tendency to annotate the entities along with its attributes, for example ``benign prostatic hyperplasia'', while the NER \& UMLS models focus more on the key words, in the previous example, it only identified ``hyperplasia'' as a disease\footnote{Please note that we present the examples in English for the reader's convenience. However, these excerpts have been translated from European Portuguese.}. Another aspect that might be hurting the NER performance is the wide variety of concepts fitted under the ``Diseases'' category (see Table \ref{tab:umls}). Further segmenting this category might yield better results, however further studies would be required to validate this hypothesis.  Apart from that, we can see that the NER \& UMLS model presents an $F_1$ score of 88.6\% and 95.5\% on the identification of Procedures and Drugs, respectively. 

Another interesting remark that can be made upon inspection of Table \ref{tab:results} is the fact that the NER models greatly benefit from the entity linking with UMLS. This is mostly due to the removal of false entities that were identified by the NER models that do not find any match on UMLS, which results in a major boost in precision. However, by comparing the NER and the NER \& UMLS models on the Procedures and Diseases datasets, one can see that the linking with UMLS also increased the recall of the NER models. This effect is due to the fact that in the NER \& UMLS model the entity type is defined by the semantic group of the entity and not by the model that identified the entity. Therefore, linking with the UMLS is also useful to classify the entity type. The same effect was not observed in the "Drugs" dataset, where the recall got worst after linking with UMLS. However,  this dataset is also the smallest so further studies would be necessary to have a better understanding of this phenomenon.

\section{Conclusion}
\label{sec:conclusion}

This work describes the methodology employed to achieve effective results for biomedical entity identification in oncology texts written in European Portuguese when no initial annotated corpus was available. That includes the annotation of a clinical corpus with the UMLS metathesaurus, the data pipeline created, and the training of three NER models for the identification of procedures, drugs, and diseases. Even though the methodology was employed for European Portuguese texts, it can be adapted to other languages with comparable resources. More precisely, the span and applicability of this methodology are dependent on three main aspects, namely: the availability of the UMLS metathesaurus for the language;\footnote{See \url{https://www.nlm.nih.gov/research/umls/knowledge_sources/metathesaurus/release/abbreviations.html\#LAT} for the complete list of languages supported by UMLS.} the availability and effectiveness of a large language model for the language; and the accessibility to clinical texts. Given these three components, the reader should be able to extrapolate this methodology to his/her project.   

\section{Future Research}
\label{sec:future}
The presented models are part of a bigger pipeline that ends up feeding a dashboard that summarizes the patient therapeutic path. Arrangements are being made so that a production test is conducted for this dashboard on the IPO emergency service. This test will expose the pipeline predictions to a panel of oncology specialists (much broader than those that developed the approach) which will not only provide a qualitative evaluation of the pipeline but also more fine-grained feedback that will allow a quantitative study. 

A path we are keen to explore is the effectiveness impact of swapping the general purpose BERTimbau \cite{Souza2020BERTimbau:Portuguese} model by the BioBERTpt \cite{Schneider2020BioBERTptRecognition} model, or even train/fine-tune BERT on European Portuguese texts has it has been shown to yield better results \cite{Unanue2017RecurrentRecognition,Newman-Griffis2018EmbeddingMobility}. This seems a natural route to further improve the performance of the system. 

\begin{acks}
This work is financed by National Funds through the Portuguese funding agency, FCT - Fundação para a Ciência e a Tecnologia, within project \grantnum{}{LA/P/0063/2020}.
\end{acks}

\bibliographystyle{ACM-Reference-Format}
\bibliography{references.bib}

\end{document}